\documentclass[runningheads]{llncs}

\usepackage{eccv}

\usepackage{eccvabbrv}
\usepackage{makecell}
\usepackage{graphicx}
\usepackage{booktabs}
\usepackage{pifont} 
\usepackage{booktabs}
\usepackage{diagbox}
\usepackage{graphicx}
\usepackage[accsupp]{axessibility}  % Improves PDF readability for those with disabilities.
\usepackage{multirow}
\usepackage{bbding}
\usepackage[pagebackref,breaklinks,colorlinks,citecolor=eccvblue]{hyperref}
\usepackage{orcidlink}

\usepackage{xcolor}
\usepackage[table]{xcolor}

\begin{document}

\title{QCA: Query- and Content-Aware Keyframe Selection for Long Video Understanding} 

\titlerunning{QCA}


\author{
Jun Peng$^\dagger$\inst{1}\orcidlink{0000-0003-0655-1594}
\and
Baiyang Song$^\dagger$\inst{1}\orcidlink{0009-0004-2901-1085}
\and
Jie Li\inst{1}\orcidlink{0000-0003-3102-6425}
\and
Hui Li\inst{1}\orcidlink{0000-0001-9139-3855}
\and
Yiyi Zhou\inst{1}\orcidlink{0000-0002-5110-4526}
\and \\
Rongrong~Ji\inst{1}\orcidlink{0000-0001-9163-2932}
\and
Yonghong Tian*\inst{2,3}\orcidlink{0000-0002-2978-5935}
}
\authorrunning{J. Peng, B. Song, and et al.}

\institute{
Key Laboratory of Multimedia Trusted Perception and Efficient Computing,\\
Ministry of Education of China, Xiamen University, P.R. China
 \and
School of Electronic and Computer Engineering, Peking University, P.R. China \and
Peng Cheng Laboratory, Shenzhen, P.R. China\\
\email{\{pengjun.cn, lijie.32\}@outlook.com, songbaiyang@stu.xmu.edu.cn,\\ \{hui, zhouyiyi, rrji\}@xmu.edu.cn, yhtian@pku.edu.cn}
}

\maketitle

\begingroup
\renewcommand{\thefootnote}{}
\footnotetext{\textsuperscript{$\dagger$}Equal contribution. 
\textsuperscript{*}Corresponding author.}
\endgroup

\begin{abstract}
Video understanding is often plagued by severe temporal redundancy, where processing dense frame sequences is both semantically inefficient and computationally expensive.
This challenge is further amplified when only a small subset of frames is truly relevant to the given query.
In this paper, we propose a Query- and Content-Aware (QCA) keyframe selection framework that can select a compact yet information-rich set of frames from long videos.
% QCA first partitions the video into temporal segments and evaluates the information contribution of each segment by jointly modeling the query-frame semantic matching degree and segment content deviation.
QCA first partitions the video into temporal segments and estimates the information contribution of each segment by jointly modeling query relevance and content deviation, and dynamically allocates keyframe budget to each segment.
Within each segment, QCA anchors on the most query-relevant frame and iteratively incorporates additional frames to maximize diversity while maintaining high semantic relevance to the query.
Crucially, our method requires no additional training and can be seamlessly integrated into existing Video-LLMs.
% Extensive experiments on multiple long video understanding benchmarks demonstrate that our proposed approach achieves better performance with fewer frames, \emph{e.g.}, $66.7\%$ on LongVideoBench with $64$ frames and $7$B model. 
Extensive experiments across multiple long video understanding benchmarks demonstrate that our proposed approach achieves state-of-the-art performance and has strong generalization ability.
% For instance, QCA achieves 67.8\% on LongVideoBench and 76.8\% on MLVU using 128 frames, while GPT-4o achieves 66.7\% and 64.6\% using 256 frames.
For instance, QCA achieves 67.8\% on LongVideoBench using 128 frames, while GPT-4o achieves 66.7\% using 256 frames.
Our codes are available in \href{https://github.com/hktk07/QCA}{GitHub}.

% \href{https://anonymous.4open.science/r/keyframe-4783}{https://anonymous.4open.science/r/keyframe-4783}.
  % \keywords{Keyframe Selection \and Video Understanding \and Training-Free}

\end{abstract}

\section{Introduction}
\label{sec:intro}

\begin{figure}[t]
  \centering
  \includegraphics[width=0.7\linewidth]{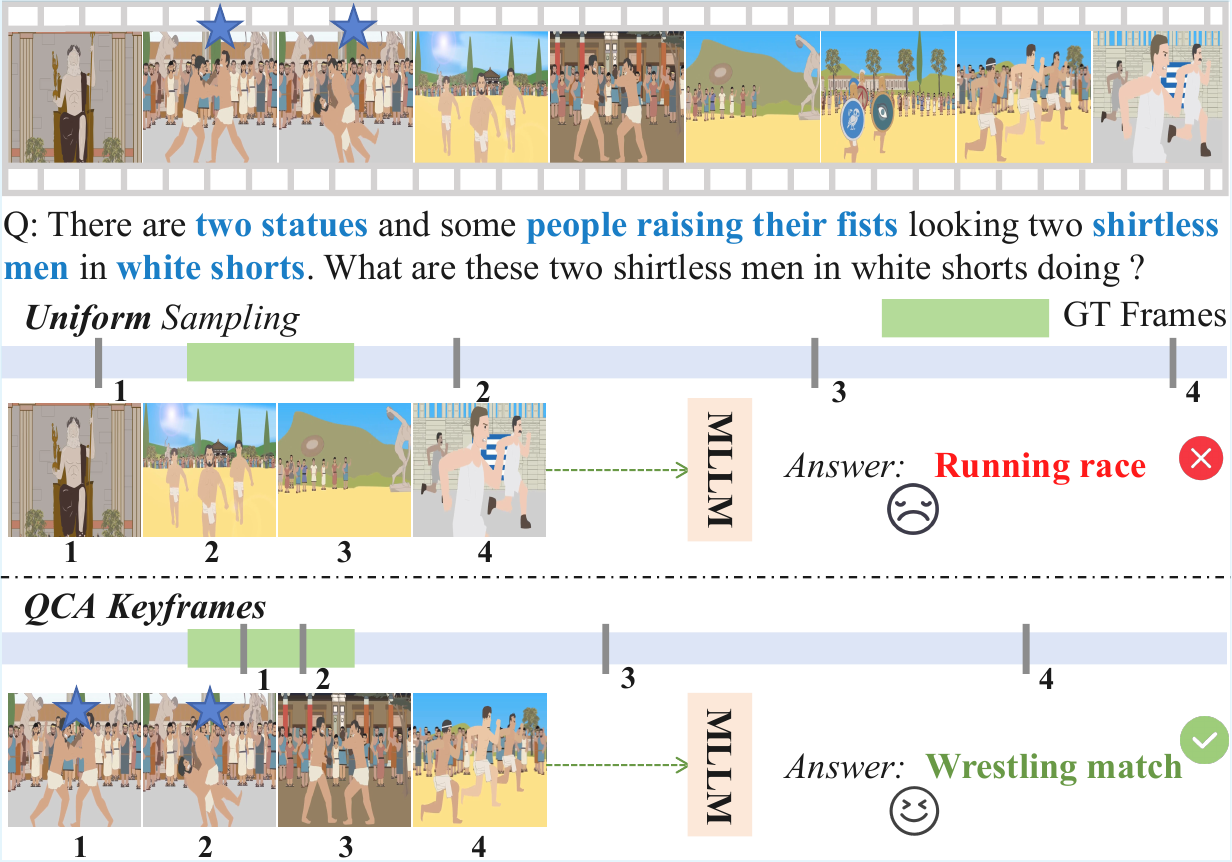}
  \caption{
  Uniform sampling may overlook semantically critical moments due to the temporal redundancy in videos, whereas our query- and content-aware selection prioritizes informative and diverse frames, enabling more reliable long video understanding under limited frame budgets.
  }
  \label{fig:intro}
\end{figure}

Large Language Models (LLMs)~\cite{brown2020language,dubey2024llama,yang2025qwen3} and Multimodal LLMs (MLLMs)~\cite{liu2023visual,Qwen3-VL,wang2025internvl3} have recently demonstrated remarkable capabilities in unified vision-language understanding, enabling complex reasoning over images and videos~\cite{li2024llava,Qwen3-VL,wang2025internvl3}.
By leveraging large-scale pre-training, these models have significantly advanced in a wide range of video understanding, \emph{e.g.}, video question answering~\cite{fu2025video}, retrieval~\cite{wu2025large}, and grounding~\cite{soldan2022mad}.

However, despite their strong representational power, applying LLMs or MLLMs to long-form video remains a challenge.
Long videos are inherently temporally redundant, and naively encoding dense frame sequences leads to substantial computational inefficiency.
More importantly, when video frames are converted into visual tokens, long videos can easily exceed the token budget limitations of current MLLMs, resulting in truncated input or degraded reasoning performance.
% Consequently, only a small subset of frames can be processed, making effective frame selection a critical problem for VLM-based video understanding.
Under such constraints, selecting a small yet informative subset of frames is often more effective than processing dense but redundant frame sequences.

Existing approaches typically address this issue through uniform sampling~\cite{zhang2024video}, attention- or relevance-based selection~\cite{dong2022identifying,arslan2023key}.
However, uniform sampling often overlooks semantically critical moments, as illustrated in Fig.\ref{fig:intro}.
Attention-based mechanisms offer fine-grained modeling, but they remain computationally expensive for long videos.
% % relevance-based filtering methods, on the other hand, focus primarily on query-frame similarity and often overlook the intrinsic structure and diversity of video content, leading to redundant or uninformative frame selections.
% And if relevance-based methods rely solely on query-frame similarity may ignore video content diversity, often resulting in redundant or insufficiently informative frame selections.
And relying solely on query-frame similarity may overlook video content diversity, leading to redundant frame selection or insufficient information.

Therefore, effective long video understanding with MLLMs requires query- and content-aware keyframe selection.
% Instead of introducing new primitives for frame scoring, our goal is to provide a principled and structured framework that jointly integrates relevance, diversity, and budget allocation for Video-LLMs.
Intuitively, different temporal segments of a video contribute unequally to a given query, and selected frames should be both semantically relevant and content-diverse within each segment.

Motivated by this insight, we propose a \textbf{Q}uery- and \textbf{C}ontent-\textbf{A}ware keyframe selection framework, which is denoted as \textbf{QCA}.
It first estimates the information contribution of each temporal segment by jointly modeling the semantic relevance and the content deviation.
Then dynamically allocate keyframe budgets across segments, followed by an intra-segment keyframe selection strategy that balances semantic alignment and content diversity.
Moreover, QCA is training-free and can be seamlessly integrated into existing Video-LLMs.

Extensive experiments on multiple long video understanding benchmarks show that our proposed QCA achieves the SOTA performance, \emph{e.g.}, $66.9\%$ on LongVideoBench~\cite{wu2024longvideobench} and $70.1\%$ on Video-MME~\cite{fu2025video}.
Besides, it achieves average performance gains of $4.00\%$ on LLaVA-Video~\cite{zhang2024video} and $4.78\%$ on Qwen3-VL~\cite{Qwen3-VL}, validating its generalization ability across different cutting-edge Video-LLMs.

% our method consistently improves performance under strict frame budgets.
% Specifically, achieving $66.7\%$ on LongVideoBench cite{}, $51.7\%$ on LVBench cite{}, which demonstrates the effectiveness and generalization of our proposed xxx.

Our main contributions are as follows:
\begin{itemize}
\item
We formulate query-conditioned keyframe selection for long video understanding as a joint relevance–diversity modeling and frame allocation problem under a limited frame budget.
\item
We propose QCA, a query- and content-aware keyframe selection framework that integrates segment-level contribution estimation, adaptive budget allocation, and anchor-centric greedy frame selection.
% \item 
% The proposed QCA is a training-free and plug-and-play keyframe selection framework tailored for MLLM-based long video understanding.
% \item
% We propose an inter-segment allocation strategy that enables adaptive keyframe budget allocation under limited frame constraints, and an intra-segment selection strategy that chooses keyframes considering both semantic matching and content diversity.
\item
Extensive experiments demonstrate consistent improvements across benchmarks, Video-LLMs, and VL embeddings, demonstrating its superiority and generalization ability.
% Extensive experiments demonstrate that QCA consistently improves performance under strict frame budgets while remaining model-agnostic to different vision–language embeddings.

\end{itemize}
% =====================================================================
\section{Related Work}
\subsection{MLLMs for Video Understanding}
Multimodal Large Language Models (MLLMs) have recently achieved significant progress in video understanding by extending LLMs with visual perception modules.
Early works such as Flamingo~\cite{alayrac2022flamingo} demonstrated the effectiveness of aligning pretrained vision encoders with frozen language models for multimodal reasoning.
Subsequent models, including LLaVA~\cite{liu2023visual} and its variants, further improved vision–language alignment through instruction tuning, achieving strong performance across a wide range of multimodal tasks.

Building on these advances, recent studies have adapted MLLMs to video understanding by representing videos as sequences of visual tokens extracted from sampled frames.
Representative approaches, such as Video-LLaMA~\cite{zhang2023video} and VideoChat~\cite{li2025videochat}, allow MLLMs to process temporal visual information together with textual queries, enabling applications including video question answering, retrieval, and captioning.
% which enable MLLMs to reason over temporal visual information and textual queries.
% These models have demonstrated promising results on video question answering, retrieval, and captioning benchmarks, highlighting the potential of MLLMs as unified frameworks for video understanding.

However, most existing Video-LLMs rely on uniform frame sampling to compress videos into a limited set of frames.
While simple and efficient, this strategy often struggles with long-form videos due to severe temporal redundancy and the limited context length of language models.
Consequently, naively increasing the number of input frames leads to a higher computational cost or truncated visual context.
% motivating the need for more efficient video representations, \emph{e.g.}, effective keyframe selection, tailored for MLLM-based video understanding.
These limitations motivate the need for more efficient video representations, such as query-aware keyframe selection.

\begin{figure*}[t]
  \includegraphics[width=1\linewidth]{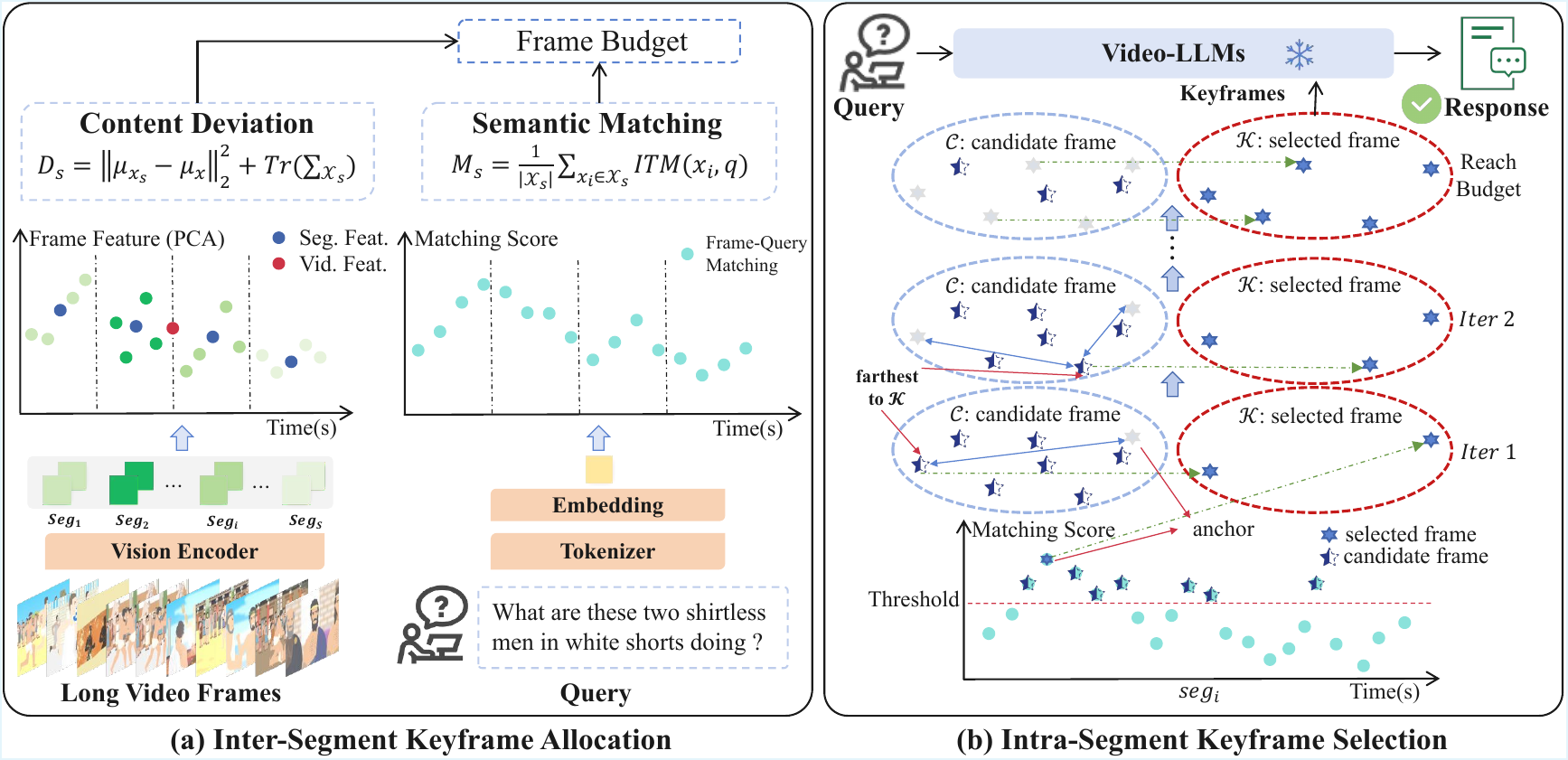} 
  \caption {
%   Overview of our proposed two-stage keyframe sampling framework. (a) Stage 1: Adaptive Budget Allocation. The long video is divided into uniform clips. We assign a frame budget to each clip based on a weighted combination of visual saliency (derived from clip distance and variance, $S^{DV}$),  and semantic relevance to the user query($S^{ITM}$). (b) Stage 2: Intra-clip Selection. Within each clip, frames are first filtered by a semantic threshold to form a candidate set $\mathcal{C}$
% . Initializing with the highest-scoring anchor frame, we employ Farthest Point Sampling to iteratively select visually diverse keyframes($\mathcal{K}$) until the assigned budget is reached.
The proposed QCA consists of: (a) Inter-Segment Keyframe Allocation, where the frame budget for each segment is dynamically determined by considering semantic alignment and visual content deviation; (b) Intra-Segment Keyframe Selection, which iteratively adds the frame with the maximum aggregate distance to the current keyframe set $\mathcal{K}_s$ from a relevance-filtered candidate set $\mathcal{C}_s$.
% which utilizes a semantic threshold to filter candidates and Farthest Point Sampling to identify the most representative and diverse frames($\mathcal{K}$) from the candidate set($\mathcal{C}$). 
 }
  \label{fig_framework}
\end{figure*}

\subsection{Keyframe Selection for Video-LLMs}
To address the scalability challenges of video inputs, extensive research has focused on designing efficient video representations for Video-LLMs. 
A common strategy is uniform sampling~\cite{zhang2024video,li2024llava}, which reduces the number of frames while maintaining coarse temporal coverage.
However, such methods are query-agnostic and often select redundant or irrelevant frames for specific queries.

In addition, some works~\cite{liang2024keyvideollm,hu2025m} have explored relevance-based frame filtering, where frames are ranked according to their semantic similarity to a given query and only the top-ranked frames are retained.
While these approaches improve query awareness, they typically overlook the intrinsic structure and diversity of video content, leading to suboptimal coverage of important temporal segments.
Clustering- and diversity-based methods~\cite{arslan2023key,zhang2025beyond,song2026ktv} have also been investigated to select representative frames, but they are usually query-independent and may fail to capture query-specific evidence.
 
More recently, several studies have proposed segment-based video representations to better handle long videos under limited frame or token budgets. 
These methods aim to summarize or compress video content before feeding it into language models, for example, through temporal segmentation~\cite{sun2025frames,zhu2025focus}, adaptive keyframe selection~\cite{tang2025adaptive,liu2025bolt,chen2026wavelet}, or token pruning~\cite{bolya2022token,chen2024image,ju2026forestprune}. 
% Despite their effectiveness, most existing approaches consider either query relevance or content diversity in isolation, and rarely integrate both aspects in a unified, query-conditioned manner.

In contrast to prior works, our approach jointly models query relevance and content structure to select a compact yet informative set of keyframes. 
By dynamically allocating a limited frame budget across temporal segments and performing semantic-anchored and diversity-aware selection within each segment, our method provides an efficient video representation that is particularly well-suited for long video understanding operating under frame budget constraints.

\section{Method}
Rather than introducing new frame-scoring primitives, our goal is to provide a structured framework that integrates query relevance, content representativeness, and budget allocation for efficient keyframe selection in long videos.
Fig.\ref{fig_framework} illustrates the overview of our proposed QCA, which first adaptively assigns the keyframe budget to different temporal segments by modeling semantic matching and content deviation.
QCA then iteratively selects keyframes from the candidate set, satisfying both semantic alignment with the query and semantic diversity between selected keyframes.

\subsection{Problem Formulation}
Given a video uniformly sampled at 1fps, the resulting frame sequences are denoted as
\begin{equation}
    \label{eq:video_set}
    \mathcal{X}=\{x_1, x_2, ..., x_N\},
\end{equation}
where $N$ is the total number of frames.
Given a query $q$, a natural language question, our goal is to select a compact keyframe subset
\begin{equation}
    \label{eq:keyframe_subset}
    \mathcal{K}\subset\mathcal{X}, |\mathcal{K}|=N',N'\ll N
\end{equation}
such that $\mathcal{K}$ preserves the most query-relevant and informative visual content for the downstream video understanding task:
\begin{equation}
    \label{eq:task}
    answer = MLLMs(\mathcal{K},q)
\end{equation}

\subsection{Temporal Segmentation}
% To capture coarse temporal structure, we uniformly partition the given video $\mathcal{X}$ into $S$ non-overlapping temporal segments:
To capture coarse temporal structure, we divide the video $\mathcal{X}$ into $S$ uniformly spaced segments:
\begin{equation}
    \label{eq:segment}
    \mathcal{X}=\bigcup_{s=1}^{S}\mathcal{X}_{s},
    \mathcal{X}_{s}\cap\mathcal{X}_{s'}=\emptyset.
\end{equation}
where $S$ is set to $12$ by default.
% This segmentation allows us to estimate the relative importance of different temporal regions before performing fine-grained keyframe selection.
Intuitively, a larger $S$ provides finer temporal granularity but reduces the number of frames allocated per segment, while a smaller $S$ may overlook short but critical events.

\subsection{Inter-Segment Keyframe Allocation}
% \textbf{Query–Frame Semantic Matching.}

\textbf{Semantic Matching.} 
We first measure the average semantic matching degree $M_s\in [0,1]$ between frames in $\mathcal{X}_s$ and the given query:
\begin{equation}
    \label{eq:itm_matching}
    M_s=\frac{1}{|\mathcal{X}_s|}\sum_{x_i\in\mathcal{X}_s}ITM(x_i,q),
\end{equation}
where $ITM(\cdot,\cdot)$ denotes Image-Text Matching function, \emph{e.g.}, BLIP-2~\cite{li2023blip}.
% This term captures how relevant a video segment is to the query at a semantic level.
This term refers to the relevant video segment to the query at the semantic level.

% \noindent
% \textbf{Segment-Video Content Deviation.} 

\textbf{Content Deviation.}
We define the segment content deviation as
\begin{equation}
    \label{eq:deviation}
    D_s=\|\mu_{\mathcal{X}_s}-\mu_{\mathcal{X}}\|_2^2 + Tr(\Sigma_{\mathcal{X}_s}),
\end{equation}
where $\mu_{\mathcal{X}_s}$ and $\Sigma_{\mathcal{X}_s}$ denote the mean feature and the covariance matrix of $\mathcal{X}_s$, respectively.
The first term measures how much a segment differs from the whole video, encouraging the selection of visually distinctive segments that are more likely to contain informative events.
The second term, trace of the covariance matrix, captures intra-segment variation, where larger values indicate rich visual diversity.
In practice, both terms are normalized to mitigate scale mismatch.

\textbf{Keyframe Allocation.}
The final information contribution score of segment $\mathcal{X}_s$ is defined as a weighted sum:
\begin{equation}
    \label{eq:contribution}
    c_s=\alpha\cdot M_s + \beta\cdot D_s,
\end{equation}
where $\alpha$ and $\beta$ are hyperparameters. And we respectively apply a softmax normalization to $M_s$ and $D_s$ across segments to ensure comparable scales before fusion.
We compute the contribution weights via
\begin{equation}
    \label{eq:contribution_weight}
    w_s = \frac{{c_s}^\tau}{\sum_{j=1}^S {c_j}^\tau},
\end{equation}
where $\tau$ controls the smoothness of allocation.
Then the number of keyframes allocated to segment $\mathcal{X}_s$ is
\begin{equation}
    \label{eq:quota}
    q_s = \lfloor w_s\cdot N'\rfloor,  \sum_{i=1}^{S} s_i = N^{'}
\end{equation}

% Since the floor operation may lead to fewer than the target number of frames. If so, we further allocate the remaining frame budget according to the segment scores. 
% If the initial allocation yields a deficit of $n$ frames, the remaining frames are assigned to the $n$ highest-scoring segments, each receiving one additional frame, ensuring that the final frame number exactly matches the target budget.
Since the floor operation may yield fewer frames than the target budget, the remaining frames are assigned to the highest-scoring segments.
Specifically, if a deficit of $n$ Specifically, if a deficit of $n$ highest-scoring segments, ensuring that the final frame count matches the target budget.

\subsection{Intra-Segment Keyframe Selection}
% Within each segment $\mathcal{X}_s$, we select $q_s$ keyframes using a semantic and diversity-driven iterative strategy.
With each segment $\mathcal{X}_s$, we anchor the most query-relevant frame and iteratively incorporate additional frames to maximize diversity while maintaining high semantic relevance.

\textbf{Semantic Anchor.}
We first select the frame with the highest relevance to the query that is added to the keyframe set $\mathcal{K}_s$
\begin{equation}
    \label{eq:anchor}
    \mathcal{K}_s = \{\arg\max_{x_i\in\mathcal{X}_s}ITM(x_i,q) \}.
\end{equation}

\textbf{Semantic Constraint.}
Let $R^*$ denote the matching score of the anchor frame. 
We construct a candidate set 
\begin{equation}
    \label{eq:candidate}
    \mathcal{C}_s = \{x_j\in\mathcal{X}_s|ITM(x_j,q)\geq\gamma\cdot R^*\},
\end{equation}
% where $\gamma$ is a threshold hyperparameter.
in which $\gamma$ controls the size of the candidate pool by filtering frames whose relevance is below a fraction of the anchor score.
Smaller $\gamma$ increases diversity but introduces noise, while larger $\gamma$ improves relevance but reduces candidate diversity.
We set $\gamma=0.7$ by default.

\textbf{Content Diversity.}
We iteratively expand $\mathcal{K}_s$ by selecting a frame that maximizes its distance to the current keyframe set:
\begin{equation}
    \label{eq:expand}
    \mathcal{K}_s = \mathcal{K}_s \cup \{\arg\max_{x_i\in\mathcal{C}_s\setminus\mathcal{K}_s}\sum_{x_j\in\mathcal{K}_s}\phi(x_i,x_j) \},
\end{equation}
where $\phi$ is a distance metric, \emph{e.g.}, Euclidean distance.
The process repeats until $|\mathcal{K}_s|=q_s$.
Notably, $\gamma$ can be reduced to broaden the range of the candidate set if $|\mathcal{K}_s|+|\mathcal{C}_s|<q_s$ initially.
The final keyframe set is obtained by aggregating selections from all segments:
\begin{equation}
    \label{eq:keyframe}
    \mathcal{K}=\bigcup_{s=1}^S\mathcal{K}_s,~|\mathcal{K}|=N'
    % \mathcal{K}=\bigcup_{s=1}^S\mathcal{K}_s
\end{equation}

The greedy strategy incrementally selects frames that maximize marginal information gain, allowing efficient approximation of diverse and representative keyframe sets.

% \input{table/stage2_code}

% \input{table/main_result}
% \input{table/sota_table}
% =====================================================================
\section{Experiments}
\subsection{Evaluation Benchmarks}
To evaluate the effectiveness of our proposed method, we conduct experiments on four widely used benchmarks for long video understanding.

\textbf{LongVideoBench}~\cite{wu2024longvideobench} highlights the referred reasoning questions that are posed on varying-length videos up to an hour long on diverse themes.
These questions are dependent on long frame input and cannot be well-addressed by a single frame or a few sparse frames. 

\textbf{Video-MME}~\cite{fu2025video} (w/o subs) spans 6 primary visual domains with 30 subfields to ensure broad scenario generalizability and encompasses both short-, medium-, and long-term videos, ranging from 11 seconds to 1 hour.

\textbf{MLVU}~\cite{zhou2025mlvu} is constructed from a wide variety of long videos, with lengths ranging from 3 minutes to 2 hours, and includes nine distinct evaluation tasks. 

\textbf{LVBench}~\cite{wang2025lvbench} comprises publicly sourced videos, including TV series, sports broadcasts, and everyday surveillance footage, and encompasses a diverse set of tasks aimed at long video comprehension and information extraction. 

\begin{table*}[t]
\centering
\caption{
% Performance comparison on LongVideoBench, Video-MME, MLVU, and LVBench. The top section presents results from state-of-the-art models using uniform sampling. The bottom sections evaluate our method against uniform sampling and recent frame selection baselines across three backbones: LLaVa-Video-7B, InternVL-3.5-8B, and Qwen3-VL-8B. $^\dagger$ denotes our re-implementation, \textbf{bold} means the highest accuracy on different MLLM, and ``-'' indicates unavailable results.
Performance comparison with existing keyframe selection methods on LongVideoBench, Video-MME, MLVU, and LVBench. We evaluate our approach against uniform sampling, Top-$k$ selection, which selects the most image-text matched frames, and recent frame selection baselines across three MLLM backbones: LLaVA-Video-7B, InternVL-3.5-8B, and Qwen3-VL-8B with 64 frames budget. $^\dagger$ denotes our re-implementation. \textbf{Bold} indicates the best performance for each backbone. ``-'' denotes unavailable results. ``*'' denotes trained.
% Improvements over the Uniform baseline are marked in \textcolor{green!80!black}{green}.
}
\label{tab:main}
\setlength{\tabcolsep}{8pt}
\resizebox{1\linewidth}{!}{%
\begin{tabular}{lccccccc}
\toprule
\multirow{2}{*}{\textbf{Method}}
% & \multirow{2}{*}[-0.9ex]
% {\shortstack{\textbf{LLM}\\\textbf{Size}}}
&\multirow{2}{*}{\makecell[c]{\textbf{LLM}\\\textbf{Size}}}
% & \multirow{2}{*}{\textbf{Training}}
&\multirow{2}{*}{\makecell[c]{\textbf{Embedding}\\\textbf{Model}}}
% & \multirow{2}{*}{\textbf{Frames}}
% & \multirow{2}{*}{\textbf{LongVideoBench}}
&\multirow{2}{*}{\makecell[c]{\textbf{LongVideo}\\\textbf{Bench}}}
% & \multirow{2}{*}[-0.9ex]
% {\shortstack{\textbf{LongVideo}\\\textbf{Bench}}}
% & \multirow{2}{*}{\textbf{Video-MME}}
&\multirow{2}{*}{\makecell[c]{\textbf{Video}\\\textbf{MME}}}
% & \multirow{2}{*}[-0.9ex]
% {\shortstack{\textbf{Video}\\\textbf{MME}}}
& \multirow{2}{*}{\textbf{MLVU}}
& \multirow{2}{*}{\textbf{LVBench}} \\
& & & & & & & \\
\midrule

\multicolumn{4}{l}{\textbf{LLaVa-Video-7B}} & \multicolumn{4}{c}{} \\

+ \textit{Uniform}
& 7B
& -
& 58.9
& 64.4
& 70.8
& 41.9 \\

+ \textit{Top-k}
& 7B
& BLIP
& 61.6
& 63.7
& 72.8
& 47.2 \\

+ \textit{AKS}~\cite{tang2025adaptive}
& 7B
& BLIP

& 62.7
& 65.3
& 71.8
& 47.6 \\

+ \textit{Q-Frame$^\dagger$}~\cite{zhang2025q}
& 7B
& LongCLIP

& 61.5
& 64.7
& 72.9
& 47.1 \\

+ \textit{OneClip-RAG}~\cite{chen2025towards}
& 7B
& {CLIP}$^{*}$

& 62.5
& 65.2
& 71.2
& - \\

+ \textit{BOLT}~\cite{liu2025bolt}
& 7B
& CLIP

& 62.2
& 64.6
& 70.3
& - \\

+ \textit{FRAG}~\cite{huang2025frag}
& 7B
& MLLM(7B)

& 60.6
& 63.7
& 69.2
& - \\

+ \textit{E-VRAG}~\cite{xu2025vrag}
& 7B
& LLM+CLIP

& \textbf{63.1}
& 65.4
& 70.2
& - \\

\rowcolor{blue!10}
+ \textit{QCA (Ours)}
& 7B
&  BLIP

& 62.9{~(\textcolor{green!80!black}{4.0$\uparrow$})}
& \textbf{66.1}{~(\textcolor{green!80!black}{1.7$\uparrow$})}
& \textbf{74.1}{~(\textcolor{green!80!black}{3.3$\uparrow$})}
& \textbf{48.9}{~(\textcolor{green!80!black}{7.0$\uparrow$})} \\

\midrule
\multicolumn{4}{l}{\textbf{InternVL-3.5-8B}} & \multicolumn{4}{c}{} \\

+ \textit{Uniform$^\dagger$}
& 8B
& -

& 61.3
& 61.9
& 69.9
& 42.8 \\

+ \textit{Top-k}
& 8B
& BLIP

& 62.5
& 61.6
& 70.4
& 48.7 \\

+ \textit{AKS$^\dagger$~\cite{tang2025adaptive}}
& 8B
&BLIP

& 62.9
& 62.8
& 70.5
& 47.9 \\

+ \textit{Q-Frame$^\dagger$}~\cite{zhang2025q}
& 8B
& LongCLIP

& 62.5
& 61.7
& 70.6
& 48.9 \\

+ \textit{OneClip-RAG$^\dagger$~\cite{chen2025towards}}
& 8B
& {CLIP}$^{*}$

& 62.1
& 61.7
& 70.2
& -\\

\rowcolor{blue!10}
+ \textit{QCA (Ours)}
& 8B
&  BLIP

& \textbf{63.5}{~(\textcolor{green!80!black}{2.2$\uparrow$})}
& \textbf{63.9}{~(\textcolor{green!80!black}{2.0$\uparrow$})}
& \textbf{71.3}{~(\textcolor{green!80!black}{1.4$\uparrow$})}
& \textbf{50.0}{~(\textcolor{green!80!black}{7.2$\uparrow$})} \\

\midrule
\multicolumn{4}{l}{\textbf{Qwen3-VL-8B}} & \multicolumn{4}{c}{} \\

+ \textit{Uniform$^\dagger$}
& 8B
& -

& 63.1
& 67.6
& 71.0
& 43.8 \\

+ \textit{Top-k}
& 8B
& BLIP

& 64.4
& 67.4
& 74.2
& 50.7 \\

+ \textit{AKS$^\dagger$~\cite{tang2025adaptive}}
& 8B
& BLIP

& 64.7
& 68.6
& 74.2
& 50.8 \\

+ \textit{Q-Frame$^\dagger$}~\cite{zhang2025q}
& 8B
& LongCLIP

& 65.8
& 67.9
& 74.7
& 50.7 \\

+ \textit{OneClip-RAG $^\dagger$}~\cite{chen2025towards}
& 8B
& CLIP$^{*}$

& 65.3
& 68.5
& 71.3
& - \\

\rowcolor{blue!10}
+ \textit{QCA (Ours)}
& 8B
&  BLIP

& \textbf{66.9}{~(\textcolor{green!80!black}{3.8$\uparrow$})}
& \textbf{69.5}{~(\textcolor{green!80!black}{1.9$\uparrow$})}
& \textbf{75.7}{~(\textcolor{green!80!black}{4.7$\uparrow$})}
& \textbf{51.8}{~(\textcolor{green!80!black}{8.0$\uparrow$})} \\

\bottomrule
\end{tabular}
}
\end{table*}
\begin{table*}[t]
\centering
\caption{Performance comparison on LongVideoBench, Video-MME, MLVU, and LVBench with state-of-the-art models using uniform sampling across two backbones: Qwen3-VL-8B and Qwen3-VL-30B-A3B-Instruct. $^\dagger$ denotes our re-implementation, \textbf{bold} means the highest accuracy on different MLLM, and ``-'' indicates unavailable results.
}
\label{tab:sota}

\setlength{\tabcolsep}{8pt}

\resizebox{1\linewidth}{!}{%
\begin{tabular}{lcccccc}
\toprule
\multirow{2}{*}{\textbf{Method}}
& \multirow{2}{*}{\makecell[c]{\textbf{LLM}\\ \textbf{Size}}}
& \multirow{2}{*}{\textbf{Frames}}
& \multirow{2}{*}{\textbf{LongVideoBench}}
& \multirow{2}{*}{\textbf{Video-MME}}
& \multirow{2}{*}{\textbf{MLVU}}
& \multirow{2}{*}{\textbf{LVBench}} \\
& & & & & & \\
\midrule

GPT-4o~\cite{hurst2024gpt}
& -
& 256 / 0.5fps
& 66.7
& 71.9
& 64.6
& - \\

Gemini-1.5-Pro~\cite{team2024gemini}
& -
& -
& 64.0
& 75.0
& -
& 33.1 \\

Qwen2.5-VL-72B~\cite{Qwen2.5-VL}
& 72B
& 1fps
& 60.7
& \textbf{73.3}
& 74.6
& 47.3 \\

Kimi-VL-16B-A3B~\cite{kimiteam2025kimivltechnicalreport}
&16B
&64
& 64.5
& 67.8
& 74.2
& - \\

LLaVA-Video-72B~\cite{zhang2024video}
&72B
&64
& 63.9
& 70.0
& 74.4
& 45.5 \\

InternVL3.5-38B~\cite{wang2025internvl3}
&38B
&64
& 65.7
& 70.9
& 77.0
& - \\

% Video-LLaVA~\cite{lin2024video}
% & 7B
% & 8
% & -
% & 39.9
% & 36.2
% & - \\

% Oryx-1.5~\cite{liu2024oryx}
% & 7B
% & 64
% & 56.3
% & 58.3
% & 67.5
% & - \\

% VideoChat2~\cite{li2023videochat}
% & 7B
% & 8
% & -
% & 39.5
% & 44.5
% & 39.3 \\

% LongVA~\cite{zhang2024longva}
% & 7B
% & 128
% & -
% & 52.4
% & 56.3
% & - \\

% Video-XL~\cite{shu2025video}
% & 7B
% & 128 / 256
% & 50.7
% & 55.5
% & 64.9
% & - \\

NVILA~\cite{liu2025nvila}
& 7B
& 256
& 57.7
& 64.0
& 70.1
& - \\

% LLaVA-OneVision~\cite{li2024llava}
% & 7B
% & 32
% & 56.4
% & 58.2
% & 64.7
% & - \\

mPLUG-Owl3~\cite{ye2024mplug}
& 8B
& 128
& 59.7
& 59.3
& 70.0
& 43.5 \\

Apollo~\cite{zohar2025apollo}
& 7B
& 2fps
& 58.5
& 61.3
& 68.7
& - \\

% LongVU~\cite{shen2024longvu}
% & 7B
% & 1fps
% & -
% & 60.9
% & 65.4
% & - \\

\midrule
% \multicolumn{3}{l}{\textbf{Qwen3-VL-8B}} & \multicolumn{4}{c}{} \\
\textbf{Qwen3-VL-8B} & 8B & 64 & 63.1&67.6&71.0&43.8\\
\rowcolor{blue!10}
+\textit{QCA (Ours)}
& 8B
& 64
% & \textbf{66.9}{~(\textcolor{green!80!black}{3.8$\uparrow$})}
% & \textbf{70.1}{~(\textcolor{green!80!black}{2.5$\uparrow$})}
% & \textbf{75.8}{~(\textcolor{green!80!black}{4.8$\uparrow$})}
% & \textbf{51.8}{~(\textcolor{green!80!black}{8.6$\uparrow$})} \\
&66.9
& 69.5
& 75.7
& 51.8 \\

\midrule
% \multicolumn{3}{l}{\textbf{Qwen3-VL-30B-A3B}} & \multicolumn{4}{c}{} \\
\textbf{Qwen3-VL-30B-A3B} & 30B & 64 & 67.2&69.9&72.8&44.0\\
\rowcolor{blue!10}
+\textit{QCA (Ours)}
& 30B
& 64
% & \textbf{66.9}{~(\textcolor{green!80!black}{3.8$\uparrow$})}
% & \textbf{70.1}{~(\textcolor{green!80!black}{2.5$\uparrow$})}
% & \textbf{75.8}{~(\textcolor{green!80!black}{4.8$\uparrow$})}
% & \textbf{51.8}{~(\textcolor{green!80!black}{8.6$\uparrow$})} \\
& \textbf{69.9}
& 71.4
& \textbf{77.1}
& \textbf{52.6} \\

\bottomrule
\end{tabular}
}
\end{table*}

\subsection{Implementation Details}
We investigate LLaVA-Video~\cite{zhang2024video}, InternVL-3.5~\cite{wang2025internvl3} and Qwen3-VL~\cite{Qwen3-VL} as our baselines. 
These MLLMs typically receive $64$ uniformly sampled video frames, which may result in the loss of key information.
For a fair comparison, we also set the number of retrained frames $N'=64$ in QCA.
Following previous research~\cite {tang2025adaptive}, we first sample the frame sequence from the raw video at $1$ FPS before performing keyframe selection to reduce computational costs.

% The Image-Text Matching (ITM) function in Eq.\ref{eq:itm_matching} is BLIP-2~\cite{li2023blip}, in which the text and visual encoders receive the query and frame to measure their matching degree. Meanwhile, for the Content Deviation term in Eq.~\ref{eq:deviation}, we extract frame-level visual features using BLIP-2~\cite{li2023blip}.

For the Image-Text Matching (ITM) function in Eq.\ref{eq:itm_matching} and the Content Deviation term in Eq.\ref{eq:deviation}, we both use BLIP-2~\cite{li2023blip} as the default embedding model, in which the text and visual encoders receive the query and frame to measure their matching degree. 

% For the Content Deviation term in Eq.~\ref{eq:deviation}, we extract frame-level visual features using DINOv2~\cite{oquabdinov2}. 
Also, we set the weighting coefficients for semantic relevance and content deviation to $\alpha=\beta=0.5$, and use a unified softmax temperature $\tau=0.5$ for all softmax operations. 
All experiments are conducted on 8$\times$A800 80G GPUs.

\begin{figure*}[t]
  \includegraphics[width=1\linewidth]{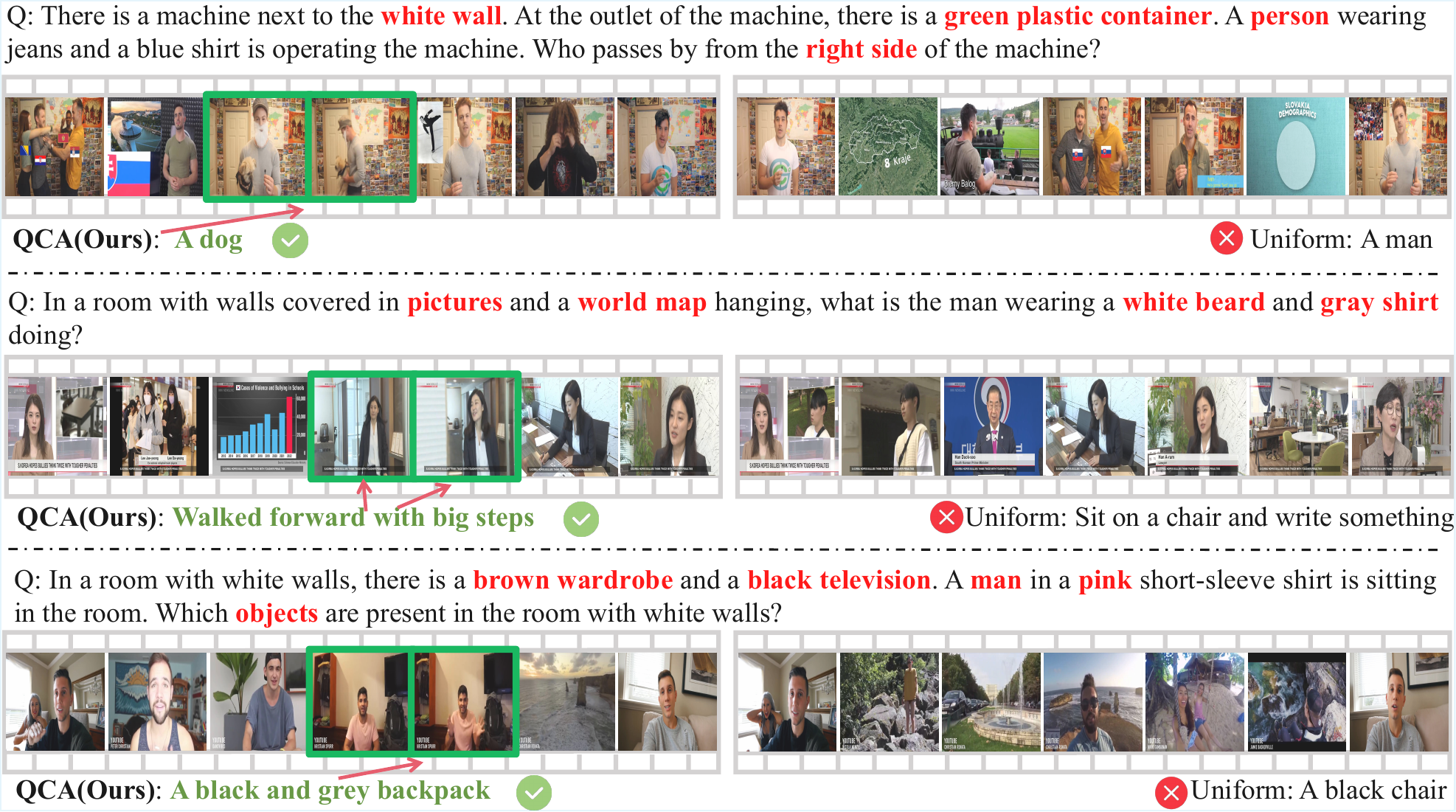} 
  \caption {Qualitative comparison of our QCA (left) and Uniform Sampling (right). The green boxes indicate the critical frames successfully selected by our method, which contain the specific evidence required to answer the questions (\emph{e.g.}, the passing dog, the white beard). In contrast, Uniform Sampling fails to capture these informative moments due to its rigid temporal intervals, leading to incorrect answers.}
  \label{visualize}
\end{figure*}

\begin{figure}[t]
  \includegraphics[width=1\linewidth]{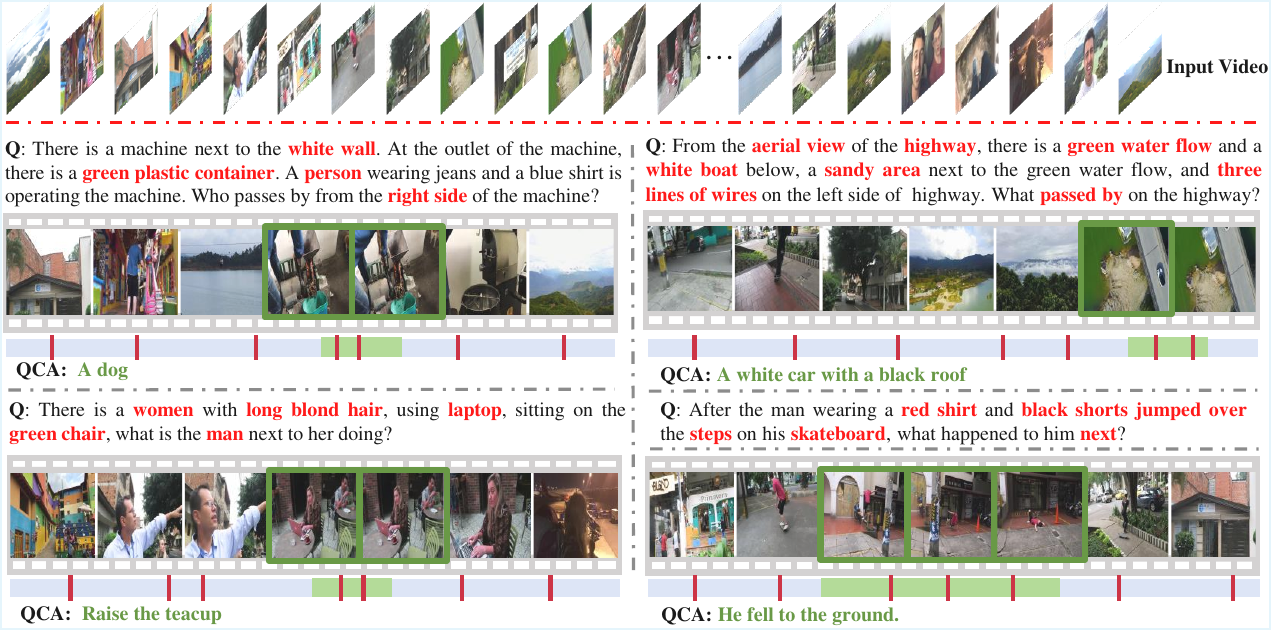} 
  \caption {For the same video, QCA selects different keyframes based on different queries, while ensuring video coverage to prevent missing key information.}
  \label{fig_different_q}
\end{figure}
\noindent

\begin{figure*}[t]
  \includegraphics[width=1\linewidth]{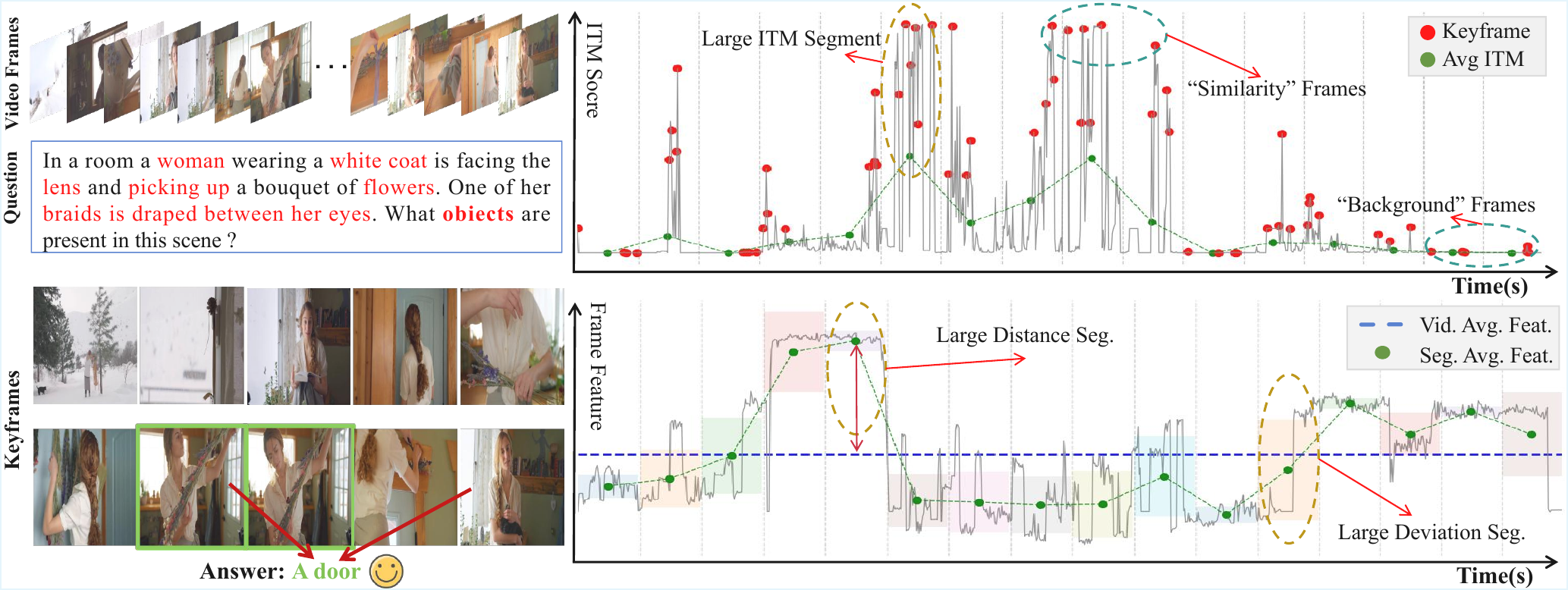} 
  \caption {An example of our QCA keyframe selection. The upper right visualizes the match score between each frame and query, while the bottom right shows the changes in frame content at the feature level. The red dot indicates the selected keyframe, and the green dot indicates the average ITM score or average frame feature within a segment. The shaded area represents the standard deviation within the segment.}
  \label{distribution}
\end{figure*}

% \subsection{Comparison to the State-of-the-Art}
% \textbf{Quantitative Results.}
\subsection{Quantitative Results}
\textbf{Comparison to existing keyframe selection methods.}
% \subsubsection{Comparison to existing keyframe selection methods}
Across four long video understanding benchmarks, Tab.\ref{tab:main} presents a comprehensive comparison between QCA and several frame selection strategies, including uniform sampling, Top-$k$ selection and recent keyframe selection methods, where Top-$k$ selects keyframes with the highest Top-$k$ image-text matching score.

Without additional training, QCA consistently achieves the best performance across representative Video-LLMs, including LLaVA-Video-7B, InternVL-3.5-8B, and Qwen3-VL-8B, demonstrating strong generalization and robustness.
In particular, with Qwen3-VL-8B under a $64$-frame budget, QCA achieves $66.9\%$ ($3.8\%\uparrow$) on LongVideoBench, $69.5\%$ ($1.9\%\uparrow$) on Video-MME, $75.7\%$ ($4.7\%\uparrow$) on MLVU, and $51.8\%$ ($8.0\%\uparrow$) on LVBench.
Although the absolute gains appear moderate, they are obtained under strict frame budgets and without additional training, making them particularly valuable for practical long video understanding scenarios.

Compared with both uniform sampling and Top-$k$ selection, QCA consistently delivers superior results across diverse tasks requiring long-range reasoning, semantic localization, and holistic video understanding.
% consistently yields substantial gains across benchmarks, whatever tasks requiring long-range reasoning, precise semantic localization or comprehensive video understanding. 
Specifically, on LongVideoBench, which emphasizes semantically critical moments, the proposed QCA significantly outperforms uniform sampling, highlighting the limitation of uniform sampling in capturing key events. 
Meanwhile, on Video-MME, which requires more comprehensive scene understanding, QCA also surpasses Top-$k$ selection.
This suggests that selecting frames solely based on high ITM scores may overly focus on locally salient moments while overlooking broader contextual information.
% These results collectively validate the effectiveness of our query- and content-aware frame selection strategy for different long video understanding in different scenarios.

We further compare QCA with recent keyframe selection methods, \emph{e.g.}, AKS~\cite{tang2025adaptive} and Q-Frame~\cite{zhang2025q}.
Our approach almost outperforms these methods across all four benchmarks. 
% In particular, QCA consistently outperforms AKS, Q-Frame which use similar VL embedding models without fine-tuning. 
% For instance, QCA gains a 2.1\% improvement over AKS on LVBench and a 2.2\% improvement against Q-Frame on Video-MME with InternVL-3.5-8B.
Specifically, QCA outperforms AKS by $2.1\%$ on LVBench and outperforms Q-Frame by $2.2\%$ on Video-MME using InternVL-3.5-8B.
Compared with FRAG~\cite{huang2025frag} and E-VRAG~\cite{xu2025vrag}, which rely on larger models for frame selection, QCA achieves comparable or superior performance. 
For instance, QCA gains a $2.4\%$ improvement on Video-MME over FRAG and a $3.9\%$ improvement on MLVU over E-VRAG using LLaVA-Video-7B.
More surprisingly, although OneClip-RAG~\cite{chen2025towards} benefits from additional training, QCA still achieves superior performance without any fine-tuning, \emph{e.g.}, gains a $2.9\%$ improvement on MLVU using LLaVA-Video-7B.
% underscoring the advantage of our structured and training-free design.
Importantly, the performance gains of QCA remain consistent across different Video-LLMs, indicating that the proposed method is model-agnostic and can be readily integrated into existing architectures.
% This consistency suggests that the observed improvements stem from better frame selection rather than model-specific characteristics.

\textbf{Comparison to state-of-art Video-LLMs.}
Tab.~\ref{tab:sota} compares QCA with state-of-the-art Video-LLMs using Qwen3-VL-30B-A3B-Instruct and Qwen3-VL-8B as backbones.
Although the Qwen3-VL series already achieve strong performance with uniform sampling of 64 frames, incorporating QCA further improves results across all benchmarks.

% For example, Qwen3-VL-30B-A3B + QCA outperforms GPT-4o by 12.5\% on MLVU and surpasses Gemini-1.5-Pro by 19.5\% on LVBench.
For example, with Qwen3-VL-30B-A3B-Instruct, QCA outperforms GPT-4o by $12.5\%$ on MLVU and surpasses Gemini-1.5-Pro by $19.5\%$ on LVBench, demonstrating that more informative frame selection can substantially enhance long video reasoning even for strong proprietary models.

Overall, these results indicate that QCA offers a simple yet effective approach for improving long video understanding by selecting a compact, informative, and diverse set of keyframes, without modifying model architectures or requiring additional training.

\subsection{Qualitative Results}
Fig.\ref{visualize} and Fig.\ref{fig_different_q} present representative video question answering examples comparing our QCA with uniform sampling. 
% Beyond selecting frames that are semantically aligned with the query, QCA also preserves globally informative context frames, even if only weakly related to the query, which helps the model maintain holistic scene understanding under a strict frame budget.
Unlike uniform sampling, which selects a fixed set of frames regardless of the query, QCA dynamically adapts keyframe selection to the question while preserving sufficient global scene coverage.

Beyond identifying frames that are strongly aligned with the query, QCA also retains context frames that provide complementary visual information, even when their direct relevance to the query is weaker. 
This helps maintain holistic scene understanding under a strict frame budget and prevents errors caused by missing contextual cues. 
% As illustrated, QCA selects keyframes that capture crucial evidence, such as the passer-by and salient objects, while still covering the overall scene structure needed for disambiguation.
As illustrated, QCA captures key evidence such as the passer-by and salient objects while still preserving the overall scene context needed for disambiguation.
% In contrast, uniform sampling often includes redundant or irrelevant content and misses key cues, leading to incorrect answers.

% Specifically, uniform sampling cannot handle different questions well for the same video because of the same sampled keyframes.
% But our QCA selects different keyframes based on different questions, while ensuring video coverage to prevent missing key information.

Fig.\ref{distribution} further visualizes the keyframe selection of QCA.
The top-right plot shows the frame-query matching score, while the bottom-right plot illustrates the content deviation at the feature level.
% As can be seen, our proposed QCA can allocate different frame budgets to each segment according to the average ITM and the average feature deviation.
Based on the average matching score and feature deviation within each segment, QCA adaptively allocates different frame budgets across segments.
% In addition, the selected keyframes satisfy semantic matching, content diversity, and temporal coverage.
The resulting keyframes jointly satisfy semantic relevance, content diversity, and temporal coverage.

% In long video understanding, the visual evidence required to answer a question is often sparse and transient—effectively a "needle in a haystack."

% As shown in the first row, the question asks about a specific event ("Who passes by..."). The relevant visual cue (the dog) appears only briefly in the video stream. Uniform sampling, which selects frames based on fixed temporal intervals, completely misses this momentary event, resulting in a hallucinated answer ("A man"). In contrast, our method leverages the semantic alignment between the query (\emph{e.g.}, "machine," "green plastic container") and the video features to locate the precise moment the event occurs, leading to the correct prediction.

% Similarly, in the third row, correctly identifying the material of the jacket ("denim" vs. "cotton") requires selecting frames with optimal lighting and proximity rather than general scene establishing shots. While uniform sampling retrieves dark or distant frames that offer little textural information, our approach adaptively selects frames that maximize information density regarding the queried entities. This demonstrates that our method successfully balances global understanding (establishing the scene context) with local detail extraction (capturing specific attributes or short actions), whereas uniform sampling often overlooks the critical frames necessary for fine-grained reasoning.

\subsection{Ablation Study}

\begin{table*}[t]
\centering
\caption{Ablation study on different components and settings. Results are reported on LongVideoBench, Video-MME, MLVU, and LVBench with Qwen3-VL.}
\label{tab:ablation_study}
% 使用 resizebox 确保表格正好占满文本宽度，如果不需要缩放可以去掉 resizebox 的壳
\resizebox{\textwidth}{!}{%
\begin{tabular}{lccccccccc}
\toprule
Setting & 
Matching & 
Deviation & 
Anchor & 
Candidate & 
Diversity & 
LongVideoBench & 
Video-MME & 
MLVU & 
LVBench 
% \multirow{2}{*}{Setting} &
% \multirow{2}{*}{Matching} &
% \multirow{2}{*}{Deviation} &
% \multirow{2}{*}{Anchor} &
% \multirow{2}{*}{Candidate} &
% \multirow{2}{*}{Diversity} &
% \multirow{2}{*}{\makecell[c]{\textbf{LongVideo}\\\textbf{Bench}}}&
% \multirow{2}{*}{\makecell[c]{\textbf{Video}\\\textbf{MME}}}&
% \multirow{2}{*}{MLVU} &
% \multirow{2}{*}{LVBench} &
\\
\midrule
Full & $\checkmark$ & $\checkmark$ & $\checkmark$ & $\checkmark$ & $\checkmark$ & \textbf{66.9} & 70.1 & \textbf{75.8} & \textbf{52.4} \\
w/o $D_s$ & $\checkmark$ & \ding{55} & $\checkmark$ & $\checkmark$ & $\checkmark$ & 66.3 & \textbf{70.4} & 75.4 & 51.0 \\
w/o $M_s$ & \ding{55} & $\checkmark$ & $\checkmark$ & $\checkmark$ & $\checkmark$ & 65.8 & 69.5 & 75.1 & 51.1 \\
w/o $D_s$, $M_s$ & \ding{55} & \ding{55} & $\checkmark$ & $\checkmark$ & $\checkmark$ & 66.0 & 68.3 & 75.2 & 51.2 \\
\midrule
w/o Anchor & $\checkmark$ & $\checkmark$ & \ding{55} (random) & \ding{55} & $\checkmark$ & 63.7 & 68.2 & 74.1 & 47.2 \\
w/o Candidate Set & $\checkmark$ & $\checkmark$ & $\checkmark$ & \ding{55} (all frames) & $\checkmark$ & 64.5 & 69.0 & 72.4 & 48.7 \\
w/o Diversity & $\checkmark$ & $\checkmark$ & $\checkmark$ & $\checkmark$ & \ding{55} (top matching) & 66.2 & 68.6 & 75.3 & 49.2 \\
% \midrule
Uniform Selection & $\checkmark$ & $\checkmark$ & \ding{55} & \ding{55} & \ding{55} & 62.7 & 67.6 & 71.8 & 43.9 \\
\bottomrule
\end{tabular}%
}
\end{table*}

\textbf{Matching and Deviation.}
In Tab.\ref{tab:ablation_study}, we first analyze the impact of semantic matching and content deviation by removing them from the full setting.
Removing either component consistently degrades performance across all benchmarks, indicating that both query relevance and content representativeness are important for effective keyframe selection.

Specifically, removing the matching term (or both terms) reduces performance from $66.9\%$ to $65.8\%$ on LongVideoBench and from $70.1\%$ to $68.3\%$ on Video-MME, highlighting the dominant role of semantic relevance in query-driven video understanding. 
Interestingly, the $2^{nd}$ row shows that using only $M_s$ still achieves $70.4\%$ on Video-MME, further confirming the importance of query-aware frame selection.

\begin{figure*}[t]
  \centering
  \includegraphics[width=1\linewidth]{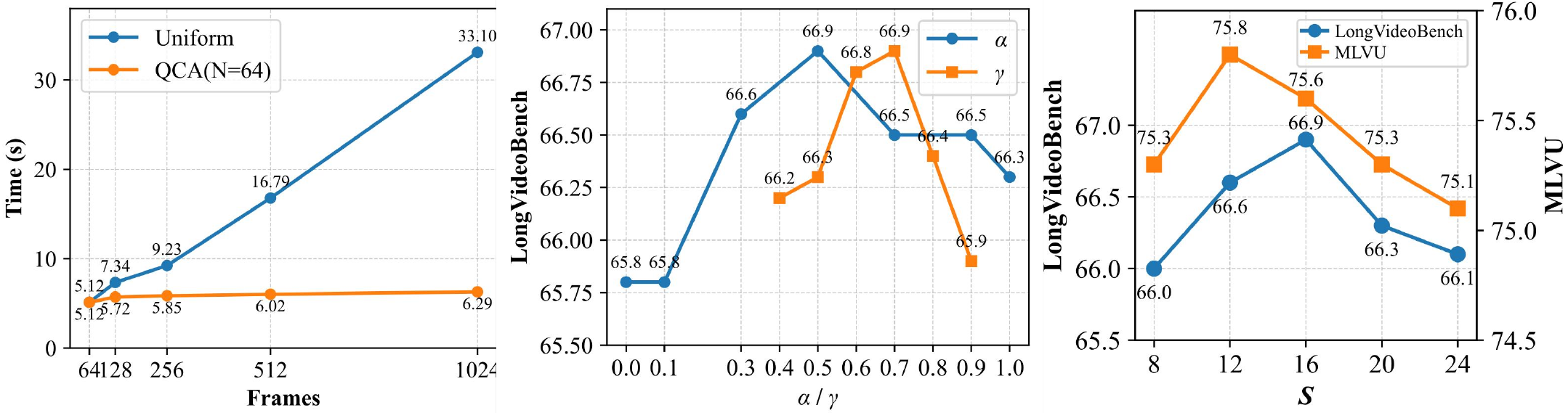}
  \caption{Time cost (left), sensitivity study in $\alpha$ and $\gamma$ (middle) and $S$(right).}
  \label{fig_time_cost_alpha_gamma_s}
\end{figure*}

Fig.\ref{fig_time_cost_alpha_gamma_s} (middle) shows the sensitivity study in $\alpha$, $\beta$ and $\gamma$, where $\beta=1-\alpha$. 
We can observe that balanced relevance-diversity weighting generally provides the best trade-off. 
Regarding $\gamma$, a smaller $\gamma$ could introduce a noisy candidate set, while a larger one leads to insufficient diversity.

% \noindent
\textbf{Anchor, Candidate, and Diversity.}
We further examine the anchor-centric selection strategy and the construction of the candidate set $\mathcal{K}_s$, as shown in the bottom of Tab.\ref{tab:ablation_study}.
When replacing the semantic anchor with a random frame, the candidate set is formed from the remaining frames in the segment. 
Performance degradation indicates that initializing the selection with the semantic anchor provides a strong semantic reference for subsequent selection.

Moreover, removing the candidate set and selecting frames from the entire segment leads to performance degradation, particularly on MLVU and LVBench. 
This result suggests that restricting the search space to high-relevance candidates helps suppress noisy frames and stabilize the diversity modeling process.

% To evaluate the need for explicit diversity modeling, we replace diversity-driven selection with a simple top-matching strategy. 
% The performance degradation on all four benchmarks suggests that diversity is essential for covering temporally dispersed yet semantically relevant events in long videos.
% In contrast, uniform selection results in substantial performance drops, confirming that our gains come from informed frame selection.
To further assess the role of diversity modeling, we replace the diversity-driven selection with a simple top-matching strategy. 
The performance drop across all benchmarks demonstrates that explicit diversity modeling is necessary to capture temporally dispersed yet semantically relevant events in long videos. 
In comparison, uniform selection causes substantial performance degradation, confirming that the improvements mainly stem from informed frame selection.

% In general, the ablation results demonstrate that each component in our framework plays a distinct and complementary role. 
% Semantic matching ensures the relevance between keyframes and the given query, deviation modeling captures content representativeness, and diversity-driven selection enables effective temporal coverage, together leading to robust performance gains across long video understanding benchmarks.
In summary, the components of QCA play complementary roles, \emph{e.g.}, semantic matching ensures query relevance, deviation modeling captures representative content, and diversity modeling improves temporal coverage.

% \begin{figure*}[t]
%   \centering
%   \includegraphics[width=1\linewidth]{seg_num_cropped (1).pdf}
%   \caption{Comparison of different $S$ used in temporal segmentation on LongVideoBench and MLVU with Qwen3-VL.}
%   \label{num_seg}
% \end{figure*}

\textbf{Temporal Segmentation.}
The number of temporal segments $S$ directly affects the effectiveness of the segment-level scoring.
As shown in Fig.~\ref{fig_time_cost_alpha_gamma_s} (right), the performance first improves and then decreases as $S$ increases. 
On LongVideoBench, the accuracy rises from $66.0\%$ at $S{=}8$ to a peak of $66.9\%$ at $S{=}16$, and then drops when $S$ becomes larger. 
A similar trend is observed on MLVU. 

When $S$ is too small, each segment covers a long temporal span and may contain multiple events, causing overly averaged statistics and inaccurate frame allocation. 
Conversely, when $S$ is too large, each segment contains fewer frames, making the statistics less stable. 
Therefore, a moderate segmentation granularity provides a better trade-off.

In addition, we explore a simple dynamic strategy in which $S$ increases linearly with video length.
However, this naive strategy did not consistently outperform using a fixed $S$.
Simply increasing the number of segments does not necessarily align with the video's underlying semantic structure. 
This indicates that QCA is relatively robust to a moderate fixed $S$. 

\textbf{Preprocessing Overhead.}
We also analyze the preprocessing overhead of QCA. 
As shown in Fig.\ref{fig_time_cost_alpha_gamma_s} (left), the keyframe selection process introduces only a marginal computational cost. 
For instance, as the number of input frames increases from 128 to 1024, the ITM matching time grows from 0.595s to 1.165s, while the selection process itself takes only about 0.005s. 
This overhead is substantially lower than the cost of processing dense visual tokens with the Video-LLM, demonstrating the efficiency of our approach.

\begin{table}[t]
    \centering
    \caption{Comparisons with different VL embedding models.}
    \label{tab:clip_abaltion}
    \setlength{\tabcolsep}{3pt}
    \small

    % \resizebox{\columnwidth}{!}{%
    \begin{tabular}{lccccc}
    \toprule
    Method & Embedding & LongVideoBench & VideoMME & MLVU & LVBench       \\ 
    \midrule
    Qwen3-VL        & -    
    & 62.7      
    & 67.6      
    & 71.0      
    & 43.8          \\ 
    \midrule
    
    +\textit{Top-k}      & \multirow{4}{*}{CLIP} 
    & 64.1      
    & 67.5      
    & 73.2      
    & 50.4          \\
    
    +\textit{Q-Frame}~\cite{zhang2025q}    &    
    & 64.7      
    & 67.8      
    & 74.1      
    & 50.6          \\
    
    +\textit{AKS}~\cite{tang2025adaptive}   &    
    & 64.3      
    & 68.5      
    & 73.7      
    & 50.1          \\
    
    \rowcolor{blue!10}
    +\textit{QCA (Ours)} &    
    & \textbf{65.5}{~(\textcolor{green!80!black}{2.8$\uparrow$})}
    & \textbf{69.4}{~(\textcolor{green!80!black}{1.8$\uparrow$})}
    & \textbf{75.3}{~(\textcolor{green!80!black}{4.3$\uparrow$})}
    & \textbf{52.4}{~(\textcolor{green!80!black}{8.6$\uparrow$})}\\ 
    \midrule
    
    +\textit{Top-k}      
    & \multirow{4}{*}{BLIP} 
    & 65.4      
    & 67.4      
    & 74.2      
    & 50.9          \\
    
    +\textit{Q-Frame}~\cite{zhang2025q}    & 
    & 65.8      
    & 67.9      
    & 74.7      
    & 50.9          \\
   
    +\textit{AKS}~\cite{tang2025adaptive}   &
    & 64.7      
    & 68.6      
    & 74.2      
    & 50.7          \\
    
    \rowcolor{blue!10}
    +\textit{QCA (Ours)} &                       
    & \textbf{66.9}{~(\textcolor{green!80!black}{4.2$\uparrow$})}
    & \textbf{70.1}{~(\textcolor{green!80!black}{2.5$\uparrow$})}
    & \textbf{75.8}{~(\textcolor{green!80!black}{4.8$\uparrow$})}
    & \textbf{51.8}{~(\textcolor{green!80!black}{8.0$\uparrow$})} \\ 
    \midrule
    
    \rowcolor{blue!10}
    +\textit{QCA (Ours)} & LongCLIP              
    & \textbf{66.2}{~(\textcolor{green!80!black}{3.5$\uparrow$})}
    & \textbf{69.5}{~(\textcolor{green!80!black}{1.9$\uparrow$})}
    & \textbf{75.6}{~(\textcolor{green!80!black}{4.6$\uparrow$})}
    & \textbf{51.9}{~(\textcolor{green!80!black}{8.1$\uparrow$})} \\ 
    \bottomrule
    \end{tabular}%
    % }

\end{table}
\textbf{VL Embeddings.}
Tab.~\ref{tab:clip_abaltion} evaluates QCA with different Vision-Language embeddings, \emph{e.g.}, CLIP, BLIP, and LongCLIP, under a fixed frame budget of $64$.
QCA consistently improves performance across different VL embedding models and benchmarks, indicating that the gains mainly stem from the proposed selection strategy rather than the choice of pre-trained encoder.
% Besides, we further evaluate QCA with LongCLIP and observe consistent improvements, confirming its robustness to the choice of encoder.
These results highlight the model-agnostic design of QCA and its ability to generalize across different VL embedding models.

% \begin{table}[t]
%     \centering
%     \caption{Comparison of different frame budgets on LongVideoBench, MLVU and LVBench with Qwen3-VL.}
%     \label{tab:frame_budget}
 
%     \resizebox{0.5\columnwidth}{!}{
%         \begin{tabular}{lcccc}
%         \toprule
%         Method & Frame & LongVideoBench & MLVU & LVBench \\
%         \midrule
%         Uniform & 64 & 63.1 & 71.0 & 43.8 \\
%         \midrule 
%         \multirow{4}{*}{QCA} & 16  & 62.7 & 71.3 & 46.3 \\
%                              & 32  & 64.5 & 73.3 & 48.9 \\
%                              & 64  & 66.9 & 75.9 & 52.4 \\
%                              & 128 & \textbf{67.8} & \textbf{76.8} & \textbf{53.2} \\
%         \bottomrule
%         \end{tabular}
%     }
% \end{table}

\begin{table}[t]
    \centering
    \caption{Comparison of different frame budgets on LongVideoBench, MLVU and LVBench with Qwen3-VL.}
    \label{tab:frame_budget}
 
    \resizebox{0.6\columnwidth}{!}{
        \begin{tabular}{lcccccc}
        \toprule
        \multirow{2}{*}{Method} & \multirow{2}{*}{Frame} &
        \multicolumn{3}{c}{LongVideoBench} &
        \multirow{2}{*}{MLVU} & \multirow{2}{*}{LVBench} \\
        \cmidrule(lr){3-5}
        & & Med & Long & Avg & & \\
        \midrule
        Uniform & 64 & 65.0 & 52.8 & 63.1 & 71.0 & 43.8 \\
        \midrule
        \multirow{4}{*}{QCA} & 16  & 63.1 & 56.6 & 62.7 & 71.3 & 46.3 \\
                             & 32  & 66.0 & 57.4 & 64.5 & 73.3 & 48.9 \\
                             & 64  & 67.2 & 59.2 & 66.9 & 75.9 & 52.4 \\
                             & 128 & \textbf{69.7} & \textbf{59.4} & \textbf{67.8} & \textbf{76.8} & \textbf{53.2} \\
        \bottomrule
        \end{tabular}
    }
\end{table}
\begin{table}[htbp]
    \centering
    \caption{Comparison with ForestPrune on VideoMME and MLVU with LLaVA-Video.}
    \label{tab:forestprune}

    \setlength{\tabcolsep}{6pt}
    \small

    % \resizebox{\columnwidth}{!}{%
    \begin{tabular}{lccccc}
    \toprule
    Method &  In Frames & Prune\(\%\) & Used Frames & VideoMME & MLVU \\
    \midrule
    LLaVA-Video & 64    & 0   & 64  & 64.4          & 70.8 \\
    ForestPrune~\cite{ju2026forestprune} & 128   & 50\% & 64 & 63.7          & 70.5 \\
    ForestPrune~\cite{ju2026forestprune} & 256   & 75\% & 64 & 64.2          & 72.5 \\
    \rowcolor{blue!10}
    QCA (Ours)  & 64    & 0   & 64  & \textbf{66.1} & \textbf{74.1} \\
    \bottomrule
    \end{tabular}%
    % }
\end{table}

% \noindent
\textbf{Keyframe Budget.} 
We further analyze the effect of different frame budgets to assess the frame efficiency of QCA.
As shown in Tab.\ref{tab:frame_budget}, our method consistently outperforms uniform sampling across all benchmarks even with only 16 or 32 frames, demonstrating that QCA can significantly reduce the number of frames or tokens required by MLLMs while maintaining strong performance.

As the frame budget increases from 16 to 128, QCA shows steady performance improvements on LongVideoBench, MLVU, and LVBench. 
This trend indicates that QCA can effectively utilize additional frames without introducing severe temporal redundancy. 
Unlike uniform sampling, which often captures redundant frames, QCA prioritizes informative and diverse keyframes, leading to more efficient use of the available frame budget.

Furthermore, we compare our method with a recent token pruning approach, ForestPrune~\cite{ju2026forestprune}, under an identical effective visual token budget. 
As shown in Tab.\ref{tab:forestprune}, QCA consistently outperforms ForestPrune on VideoMME and MLVU. 
This is likely because QCA proactively selects semantically informative frames before encoding, leading to higher information density within the fixed budget, whereas token pruning operates post-encoding and may discard tokens from less-relevant frames.

Overall, these results demonstrate a favorable trade-off between frame budget and performance, making QCA well-suited for long video understanding under strict computational constraints.

% =====================================================================
\section{Conclusion}
In this work, we study the problem of efficient long video understanding with MLLMs under a strict frame budget. 
We identify temporal redundancy as a key factor in existing Video-LLMs and propose a query- and content-aware framework that dynamically allocates frame budgets across video segments before selecting keyframes. 
By jointly modeling semantic relevance, content deviation, and diversity among selected frames, our method selects a compact yet informative set of keyframes for subsequent reasoning.
Extensive experiments demonstrate that our approach consistently outperforms uniform sampling and recent frame selection baselines across different backbones. 
In particular, our method achieves strong performance even with substantially fewer frames, highlighting its superior frame efficiency and scalability. 
% We believe this work offers a practical and model-agnostic solution for improving long video understanding and provides a promising direction for integrating structured frame selection with MLLMs.
Notably, we adopt uniform temporal segmentation for simplicity and efficiency. 
Incorporating content-aware segmentation, \emph{e.g.}, shot boundary detection, is a promising direction for future work.

\section*{Acknowledgments}
This work was supported by the New Generation Artificial Intelligence-National Science and Technology Major Project (No. 2025ZD0122701), the National Natural Science Foundation of China under Grant (No.U25B6003 and No.62425101), and the Shenzhen Science and Technology Program under Grant (No.KQTD2024
0729102051063).

\bibliographystyle{splncs04}
\bibliography{main}
\end{document}